\documentclass[10pt,twocolumn,letterpaper]{article}

\usepackage{cvpr}
\usepackage{times}
\usepackage{epsfig}
\usepackage{graphicx}
\usepackage{caption}
\usepackage{cite}
\usepackage{makecell}
\usepackage{threeparttable}
\usepackage{subfig}
\usepackage{mathrsfs}
\usepackage{booktabs}
\usepackage{multirow}
\usepackage{setspace}
\usepackage{float}
\usepackage{amsmath,amssymb,amsfonts}
\usepackage{algorithmic}
\usepackage{textcomp}

\usepackage[breaklinks=true,bookmarks=false]{hyperref}

\cvprfinalcopy % *** Uncomment this line for the final submission

\def\cvprPaperID{****} % *** Enter the CVPR Paper ID here

\begin{document}

%%%%%%%%% TITLE
\title{HCNet: Hierarchical Context Network for Semantic Segmentation}

% \Author{Congchong Nie $^{1}$, Yulong Tao $^{1}$, Xiaoshu Chen $^{1}$ }

% \address{%
% $^{1}$ \quad State Key Laboratory of Information Engineering in Surveying, Mapping and Remote Sensing, Wuhan University, Wuhan, China; ywchong@whu.edu.cn (Y.C.); nick\_ncc@whu.edu.cn (C.N.); taoyl1996@whu.edu.cn (Y.T.); xschen@whu.edu.cn (X.C.) \\}

\author{Yanwen Chong$^{1}$, Congchong Nie$^{1}$, Yulong Tao $^{1}$, Xiaoshu Chen$^{1}$, Shaoming Pan$^{1}$\\
$^{1}$State Key Laboratory of Information Engineering in Surveying, Mapping and Remote Sensing, \\
Wuhan University, Wuhan, China\\
{\tt\small nick\_ncc@whu.edu.cn}
% \and
}

\maketitle
%\thispagestyle{empty}

%%%%%%%%% ABSTRACT
\begin{abstract}
   Global context information is vital in visual understanding problems, especially in pixel-level semantic segmentation. The mainstream methods adopt the self-attention mechanism to model global context information.  However, pixels belonging to different classes usually have weak feature correlation. Modeling the global pixel-level correlation matrix indiscriminately is extremely redundant in the self-attention mechanism. In order to solve the above problem, we propose a hierarchical context network to differentially model homogeneous pixels with strong correlations and heterogeneous pixels with weak correlations. Specifically, we first propose a multi-scale guided pre-segmentation module to divide the entire feature map into different classed-based homogeneous regions.  Within each homogeneous region, we design the pixel context module to capture pixel-level correlations. Subsequently, different from the self-attention mechanism that still models weak heterogeneous correlations in a dense pixel-level manner, the region context module is proposed to model sparse region-level dependencies using a unified representation of each region. Through aggregating fine-grained pixel context features and coarse-grained region context features, our proposed network can not only hierarchically model global context information but also harvest multi-granularity representations to more robustly identify multi-scale objects. We evaluate our approach on Cityscapes and the ISPRS Vaihingen dataset. Without Bells or Whistles, our approach realizes a mean IoU of 82.8\% and overall accuracy of 91.4\% on Cityscapes and ISPRS Vaihingen test set, achieving state-of-the-art results.
\end{abstract}

%%%%%%%%% BODY TEXT
\section{Introduction}

Semantic segmentation is a vital part of the visual understanding system. It aims to parse images through assigning a class label to each pixel for an image. Currently, semantic segmentation technology has been widely used in various fields such as automatic driving \cite{Cao_2019_CVPR,Yu_2018_ECCV}, and remote sensing image interpreting \cite{marmanis2018classification,lu2019multi-scale}.

\begin{figure*}[t]
\setlength{\abovecaptionskip}{0.cm}
\setlength{\belowcaptionskip}{-0.cm}
\begin{center}
   \includegraphics[width=17cm]{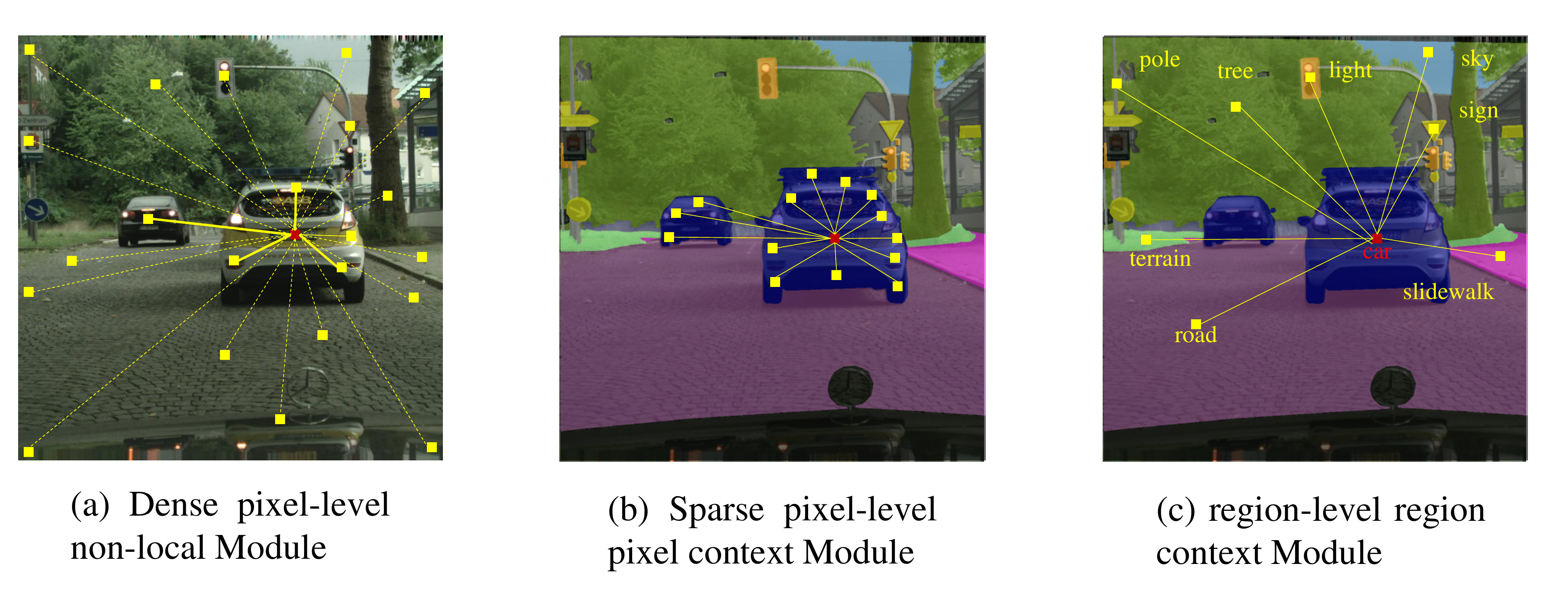}
\end{center}
   \caption{Comparison between self-attention and our proposed hierarchical context modules. The red square represents the current position, and the yellow square represents the remaining positions. The dotted line in (a) indicates weak correlation between pixels. It can be seen that the non-local module models the dense pixel-level correlation between the current position and all other positions indiscriminately. But our method first models the sparse pixel-level context between pixels of the same class. Then, the regional context between different classes is captured through the proposed region context module.}
\label{introduction1}
\end{figure*}

Traditional methods mostly adopt machine learning algorithms to perform image segmentation with various techniques, such as thresholding \cite{yuzhi}, region growing \cite{grow1}, edge detection \cite{edge1, edge2}, clustering \cite{julei1,julei2}, etc. Most successful works are based on hand-crafted features, such as HOG \cite{hog}, SIFT \cite{sift}, etc. However, with the rise of deep learning, traditional methods relying on feature engineering have gradually been replaced by the convolutional neural network (CNN) with adaptive feature learning. Block-based semantic segmentation is an early representative method based on CNN. This method first extracts regular blocks from the image in a sliding window and performs classification using common CNN (such as AlexNet \cite{alexnet}, VGG \cite{VGG}, GoogLeNet \cite{googlenet} and ResNet \cite{ResNet}). The prediction result of the image block is regarded as the class of the center pixel.  For example, Sakrapee et al. \cite{block} exploit CNN for semantic pixel labelling by cropping multi-resolution image blocks. However, the block-based methods have great redundancy due to repeated feature extraction in overlapping regions.

Things changed thoroughly after the emergence of fully convolutional network (FCN). It learns a mapping from pixels to pixels, without extracting image blocks. However, due to the fixed geometric structures, the conventional FCN is inherently limited to local receptive fields. The limitation of insufficient global context information imposes a great adverse effect on its segmentation accuracy. To make up for the above deficiency of FCN, some works obtain global context information from the perspective of multi-scale aggregation. Multiple studies \cite{FCN, UNet, RefineNet, PSPNet} adopt the pooling operation to generate multi-resolution features, which are then up-sampled and aggregated for prediction. In addition, other works \cite{DeepLabV3, DeepLabV3plus} apply dilated convolution with diverse dilated rates to acquire multi-scale contextual information. However, The above method based on multi-scale aggregation adopts a non-adaptive extraction process for all pixels, which cannot meet the requirements of different context dependencies for specific positions.

Recently, some works have focused on using the self-attention mechanism \cite{Attention_is_All_you_Need} to capture global context information for semantic segmentation. OCNet \cite{OCNet} aggregates objects context by computing the correlations of each pixel and all the other pixels. Similarly, DANet\cite{DANet} and Relational Context-aware Network \cite{zhuxiaoxiang} explore dense pixel-level contextual correlations through the self-attention mechanism in both spatial and channel dimensions. However, we found that the correlation between pixels belonging to different classes is usually weak in these methods, which means that these low correlation positions have minimal impact on the feature representation of the current position. Therefore, performing dense pixel-level modeling between these pixels will give rise to enormous redundant computation.

\begin{figure*}[htbp]
\setlength{\abovecaptionskip}{0.cm}
\setlength{\belowcaptionskip}{-0.cm}
\centering
\includegraphics[width=17cm]{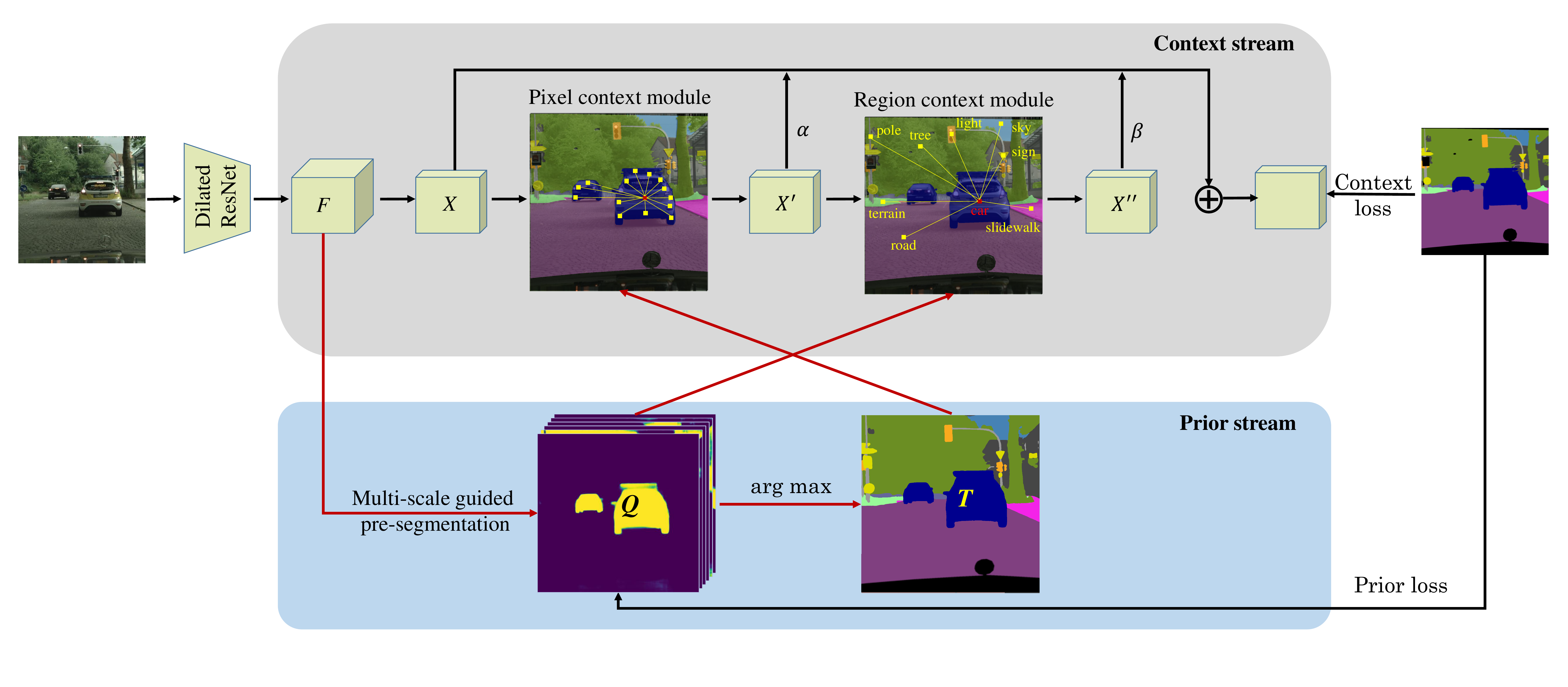}
\caption{The main structure of Hierarchical Context Network, which consists of two streams. The upper context stream is designed to hierarchically capture global context information, and the lower prior stream provides prior information of regions for the context stream.}
\label{overall_structure}
\end{figure*}

To address the drawback of the self-attention mechanism, we propose a hierarchical context network (HCNet) to model global context information. Specifically, pixel-level correlation is still captured between pixels of the same class with strong correlation, and a unified region-level correlation is modeled for heterogeneous pixels with weak correlations. As illustrated in Figure \ref{overall_structure}, we append two streams named context stream and prior stream at the end of dilated ResNet. The prior stream is designed to provide region partition result to the context stream by the proposed multi-scale guided pre-segmentation. The context stream consists of a pixel context module (PCM) and a region context module (RCM). Concretely, the PCM is first proposed for modeling pixel-level correlation between any two positions within each homogeneous region, as illustrated in Figure \ref{introduction1} (b). Subsequently, instead of performing dense pixel-level modeling between different homogeneous regions in self-attention mechanism, we capture the correlation between the region representations by proposed RCM, as illustrated in Figure \ref{introduction1} (c). The region representation is obtained by the proposed region pooling, and the enhanced region representation is restored to the pixel representation by region unpooling. Finally, we aggregate the output of the above hierarchical context modules to obtain features with global representation. In summary, our main contributions are three-fold:

$\bullet$ In order to improve the heterogeneous pixel redundancy modeling of the self-attention mechanism, We designed a HCNet to efficiently capture global context information for more accurate semantic segmentation.

$\bullet$ A PCM is proposed to learn pixel-level dependencies within each homogeneous region generated by proposed prior pre-segmentation. A RCM is designed to model region-level context between different regions with the help of the proposed region pooling and unpooling. Through aggregating fine-grained pixel context features and coarse-grained region context features, HCNet can harvest multi-granularity representations to more robustly identify multi-scale objects.

$\bullet$ The proposed HCNet achieves leading performance on two authoritative segmentation datasets used for autonomous driving and aerial interpretation, including Cityscapes and ISPRS Vaihingen datasets.

\section{Related work}
\noindent
\textbf{Multi-scale context for segmentation.} 
Fully Convolutional Networks (FCNs) \cite{FCN} successfully transform semantic segmentation into a per-pixel labeling task by replacing fully connected layers in DCNN \cite{alexnet, ResNet, VGG, googlenet, Densenet} with convolutional ones. Following that, several FCN-based works have been proposed to capture rich contextual information from the perspective of multi-scale aggregation. RefineNet \cite{RefineNet}, ExFuse \cite{ExFuse} and CCL \cite{CCL} fuse multi-resolution features through an encoder-decoder structure, which achieves the complementation of detail information and semantic information as well as obtains rich multi-scale context. Correspondingly, PSPNet \cite{PSPNet} and Deeplabv3 series \cite{DeepLabV3, DeepLabV3plus} possess abundant contextual information using parallel multi-scale branches with different sizes of pooling kernel or dilated convolution with diverse dilated rates.

\noindent
\textbf{Self-attention for segmentation.} 
The Self-attention mechanism, which is first proposed in machine translation \cite{Attention_is_All_you_Need}, has been widely used to re-model feature space according to pixel-level dependencies between each pair of pixels in computer vision. \cite{Non_Local} proposes a self-attention module to model dependencies in the space-time dimension. Due to the outstanding performance on capturing contextual information, the self-attention module has been increasingly applied in various computer vision tasks \cite{Non_Local, DANet, Detection}. OCNet \cite{OCNet} aggregates objects context by computing the similarities of each pixel and all the other pixels, which is essentially equivalent to the self-attention module. Similarly, DANet \cite{DANet} and Relational Context-aware Network \cite{zhuxiaoxiang} explore contextual dependencies through the self-attention module in both spatial and channel dimensions.

Considering that pixel-level similarities between different classes are commonly insignificant, establishing dense pixel-level dependencies leads to massive redundant relationships and high complexity in time and space. Accordingly, our proposed method captures the dependencies between pixels within each homogeneous region and continues to model the correlation between different regions. It can hierarchically capture global context information while effectively reducing the computational complexity.

\noindent
\textbf{Hierarchical structure.} 
There are a lot of successful applications of hierarchical structure, such as document classification \cite{Document}, response generation \cite{response} and action recognition \cite{Skeleton}. A hierarchical structure is designed to extract contextual information from both word and sentence levels for document classification and response generation \cite{Document, response}. In addition, for action recognition task, \cite{Skeleton} divides the human skeleton into five parts according to the physical structure of the human body, and then extracts their features respectively, which is fused to produce a final representation of the skeleton sequence at a higher hierarchy.

Our method introduces the hierarchical structure into the semantic segmentation task for the first time, in which we partition the whole feature map into several class-based homogeneous regions and then explore contextual information within and between regions from pixel-level and region-level in a hierarchical structure.

\section{Hierarchical Context Network}

\subsection{Overview}
The overall architecture of our proposed network is shown in Figure \ref{overall_structure}, which consists of a context stream and a prior stream. To begin with, an input image is processed by dilated ResNet \cite{dilated_resnet} pre-trained on ImageNet dataset \cite{ImageNet} to produce a feature map $F$ with the spatial size of $H \times W$. Considering the importance of global context, we further introduce two hierarchical context modules on the top of dilated ResNet in the context stream, including PCM and RCM, to hierarchically capture global context information from pixel-level and region-level. Meanwhile, the prior stream is designed to provide region prior information for the context stream, in which we conceived a multi-scale guided pre-segmentation strategy to partition the feature map into several class-based homogeneous regions. Finally, the feature map enhanced by global information is up-sampled to the original resolution, which is then fed into the softmax function to obtain the probability of each pixel belonging to each class. The class with the highest activation probability is considered as the final prediction of the pixel.

\subsection{Multi-scale guided pre-segmentation}

\begin{figure}[t]
\setlength{\abovecaptionskip}{0.cm}
\setlength{\belowcaptionskip}{-0.cm}
\begin{center}
  \includegraphics[width=8cm]{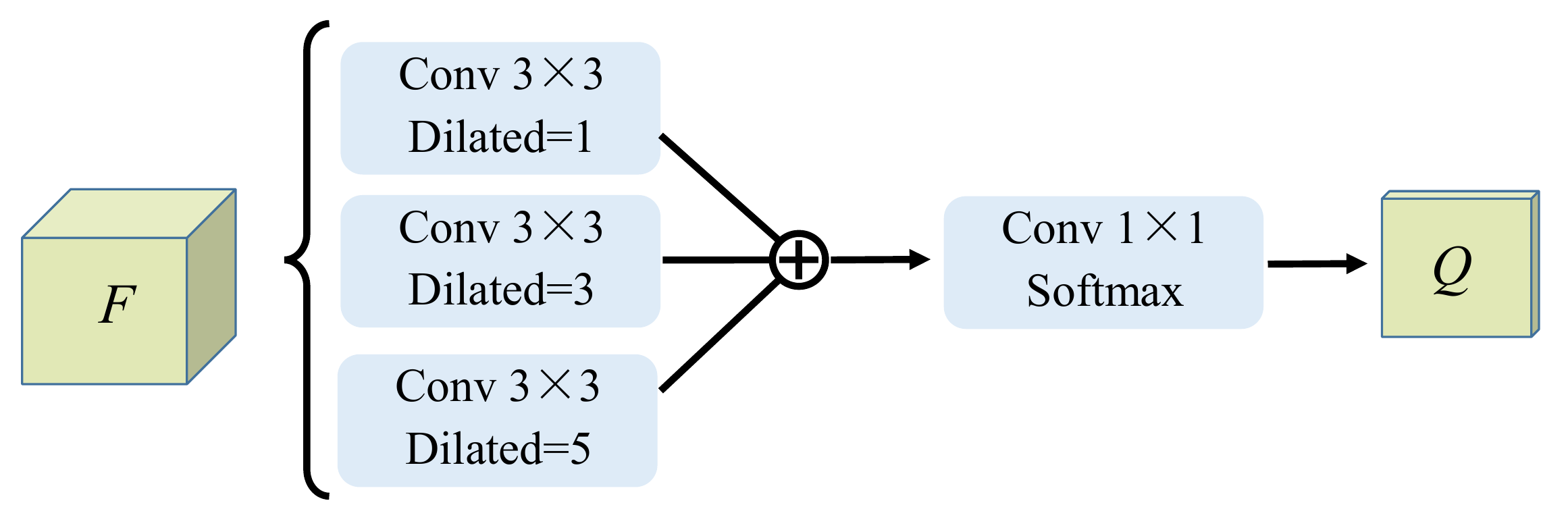}
\end{center}
  \caption{Multi-scale guided pre-segmentation module for obtaining homogeneous regions.}
\label{preseg}
\end{figure}

The prior stream aims to partition the input feature map and provide the region partition result for the context stream.
At first, we tried to use superpixel to achieve that. However, most of superpixel segmentation methods perform unsupervised iterative clustering. On the one hand, the iterative process leads to huge computational complexity. On the other hand, these methods cannot guarantee accurate semantic homogeneous regions considering the difference in object appearance and unsupervised optimization process.

Therefore, we propose a multi-scale guided pre-segmentation module, which can flexibly partition features into class-based homogeneous regions according to the supervised guidance of ground truth. As shown in Figure \ref{preseg}, the input feature map $F$ is fed into three parallel dilated convolutions with dilation rates (1, 3, 5). Each convolution output feature map has 64 channels Then, the feature maps of the three branches are aggregated through element-wise addition. Finally, a $1 \times 1$ convolution layers and softmax function is applied to obtain affiliated probability prediction $Q \in \mathbb{R}^{N \times H \times W}$, in which $N$ represents the number of classes. During training, we use prior loss $L_{prior}$ to supervise affiliated probability prediction $Q$.

The proposed multi-scale guided pre-segmentation can generate semantic homogeneous regions by introducing few convolution parameters. In particular, the convolutions with three different dilated rates can integrate multi-scale features to enhance the sensitivity of proposed pre-segmentation module to multi-scale objects. Moreover, the auxiliary supervision $L_{prior}$ can directly transfer the gradient to the shallower layer while accelerating the network training process.

\subsection{Context stream}

\subsubsection{Pixel Context module}

\begin{figure}[t]
\setlength{\abovecaptionskip}{0.cm}
\setlength{\belowcaptionskip}{-0.cm}
\begin{center}
   \includegraphics[width=8cm]{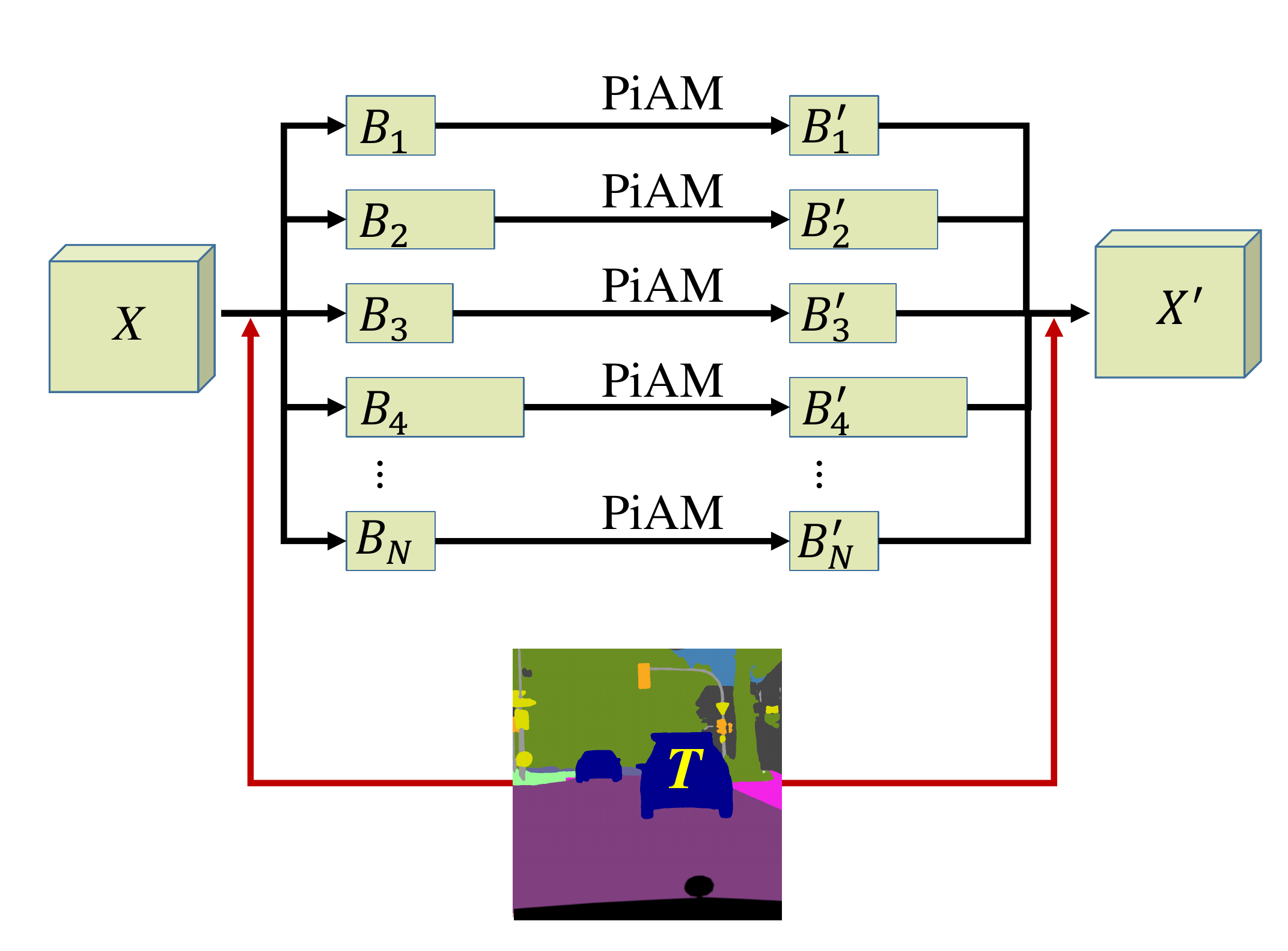}
\end{center}
   \caption{llustration of the proposed Pixel Context module. The red arrow represents the prior information guidance.}
\label{intra}
\end{figure}

Establishing pixel-level dependencies can capture rich contextual information to enhance the representation of features. Different from \cite{DANet} \cite{Non_Local} modeling dense pixel-level dependencies on the entire feature map, we introduce a relatively sparse PCM to establish pixel-level dependencies within each homogeneous region.

As illustrated in Figure \ref{intra}, taking the affiliated probability prediction $Q \in \mathbb{R}^{N \times H \times W}$ obtained by multi-scale guided pre-segmentation, we first perform $arg max$ to get explicit region boundary prediction $T \in \mathbb{R}^{H \times W}$. And then divide the input feature map $X$ into $N$ several homogeneous regions $\{B_i|1,2,...,N\}$. For each region $B_i \in \mathbb{R}^{C \times K_{i}}$, in which $K_i$ represents the number of pixels of the $i^{th}$ 
region, we capture pixel-level dependencies using Position-independent Attention Module (PiAM, detailed description below) to generate $B_{i}^{'} \in \mathbb{R}^{C \times K_{i}}$,  and $\sum_{i=1}^{N} K_{i}=H \times W$. Finally, we aggregate features $\{B_{i}^{'}|1,2,...,N\}$ to reconstruct a new feature map $X^{'} \in \mathbb{R}^{C \times H \times W}$ according to the explicit region boundary prediction $T$.

Quantitatively, given the number of regions $N$, the time complexity and spatial complexity of PCM are both $O(\sum_{i=1}^{N} K_{i}^{2}C^2)$. In particular, the complexity takes the minimum value $O(\frac{H^{2}W^{2}C^{2}}{N})$ with $K_{i}=\frac{HW}{N}$ for each block. With the help of explicit region boundary prediction $T$, our PCM models pixel-level context within the class-based homogeneous region. On the one hand, it can aggregate the features of strongly associated positions in a more sparse way to enhance the pixel representation. On the other hand, ignoring the feature of weakly associated positions at the pixel-level can improve the redundancy of self-attention while not affecting the model performance.

\subsubsection{Region Context Module}

\begin{figure}[t]
\setlength{\abovecaptionskip}{0.cm}
\setlength{\belowcaptionskip}{-0.cm}
\begin{center}
   \includegraphics[width=8cm]{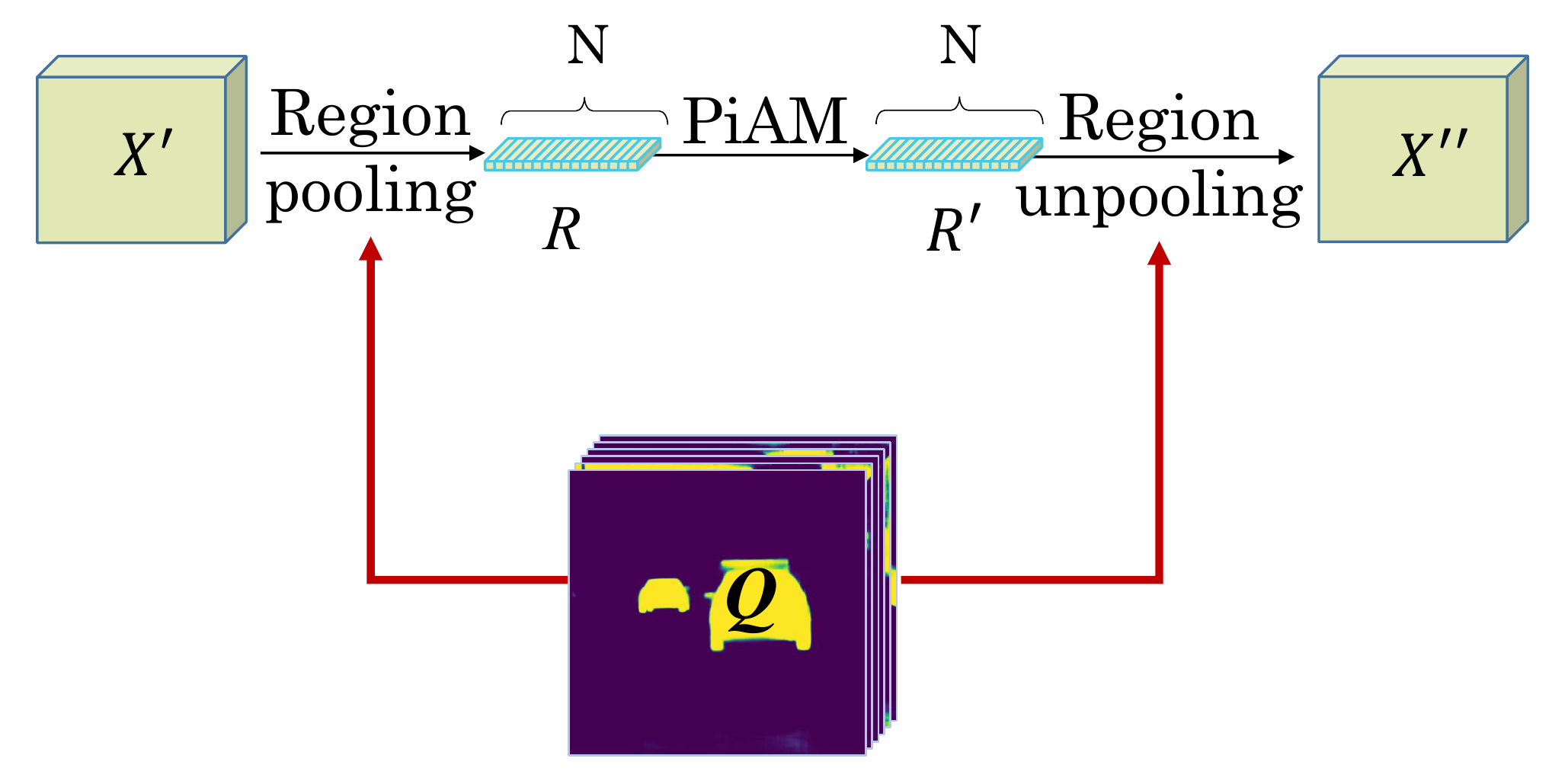}
\end{center}
   \caption{llustration of the proposed Region Context Module.}
\label{inter}
\end{figure}

The PCM only obtains the context between pixels within each homogeneous region. This section further proposes an RCM to capture the context between different regions. By combining the proposed hierarchical PCM and RCM, the global context can be completely constructed while avoiding redundant self-attention methods. As shown in Figure \ref{inter}, our RCM mainly includes region pooling, region-level attention and region unpooling.

The purpose of region pooling is to achieve the scale conversion from fine-grained pixel representation to representative region representation.
Considering that the explicit region boundary prediction $T$ cannot provide sufficient information about the relationship between pixels and each superpixels, we adopt the affiliated probability prediction $Q$ as the mapping index from pixel to region. To be more specific, given the input feature map $X^{'} \in \mathbb{R}^{C \times H \times W}$ and affiliated probability prediction $Q \in \mathbb{R}^{N \times H \times W}$, we reshape them to $\mathbb{R}^{C \times HW}$ and $\mathbb{R}^{N \times HW}$, respectively. Then the 
region representation $R \in \mathbb{R}^{C \times N}$ can be calculated as follows:
\begin{equation}
R_{i,j}= \frac{X^{'}_{i,:} Q_{j,:}^{T}}{ \sum_{k=0}^{HW}Q_{j,k}},
\end{equation}
\noindent where $R_{i,j}$ represents the feature of $i^{th}$ channel in the $j^{th}$ region. After getting the region representation, we apply the proposed PiAM to adaptively capture the region correlations and enhance region representations. Specifically, we feed the region representations $R$ into PiAM to obtain new region features $R^{'} \in \mathbb{R}^{C \times N}$. Finally, the region unpooling directly performs matrix multiplication between region features $R^{'}$ and affiliated probability prediction $Q \in \mathbb{R}^{N \times H \times W}$ to recover pixel representation $X^{''}\in \mathbb{R}^{C \times H \times W}$ as follows:
\begin{equation}
X_{i,j,k}^{''}=R_{i,:}^{'} Q_{:,j,k},
\end{equation}
\noindent where $X_{ijk}^{''}$ represents the feature of $i^{th}$ channel in row $j$ and column $k$ of feature map $X^{''}$.

Previous pooling operations usually aggregate features of regular regions indiscriminately, while our region pooling selectively aggregates pixel features according to the pixel-region affiliated probability. Therefore, it can effectively deal with irregular region pooling, and the resulting coarse-grained region features are more representative. Following that, region-level attention can adaptively capture region-level dependencies and update region  representations accordingly. The most important part is that the region unpooling can effectively recover the pixel representation with more accurate details through pixel-region affiliated probability.

\subsubsection{Position-independent attention module}

\begin{figure}[t]
\setlength{\abovecaptionskip}{0.cm}
\setlength{\belowcaptionskip}{-0.cm}
\begin{center}
   \includegraphics[width=8cm]{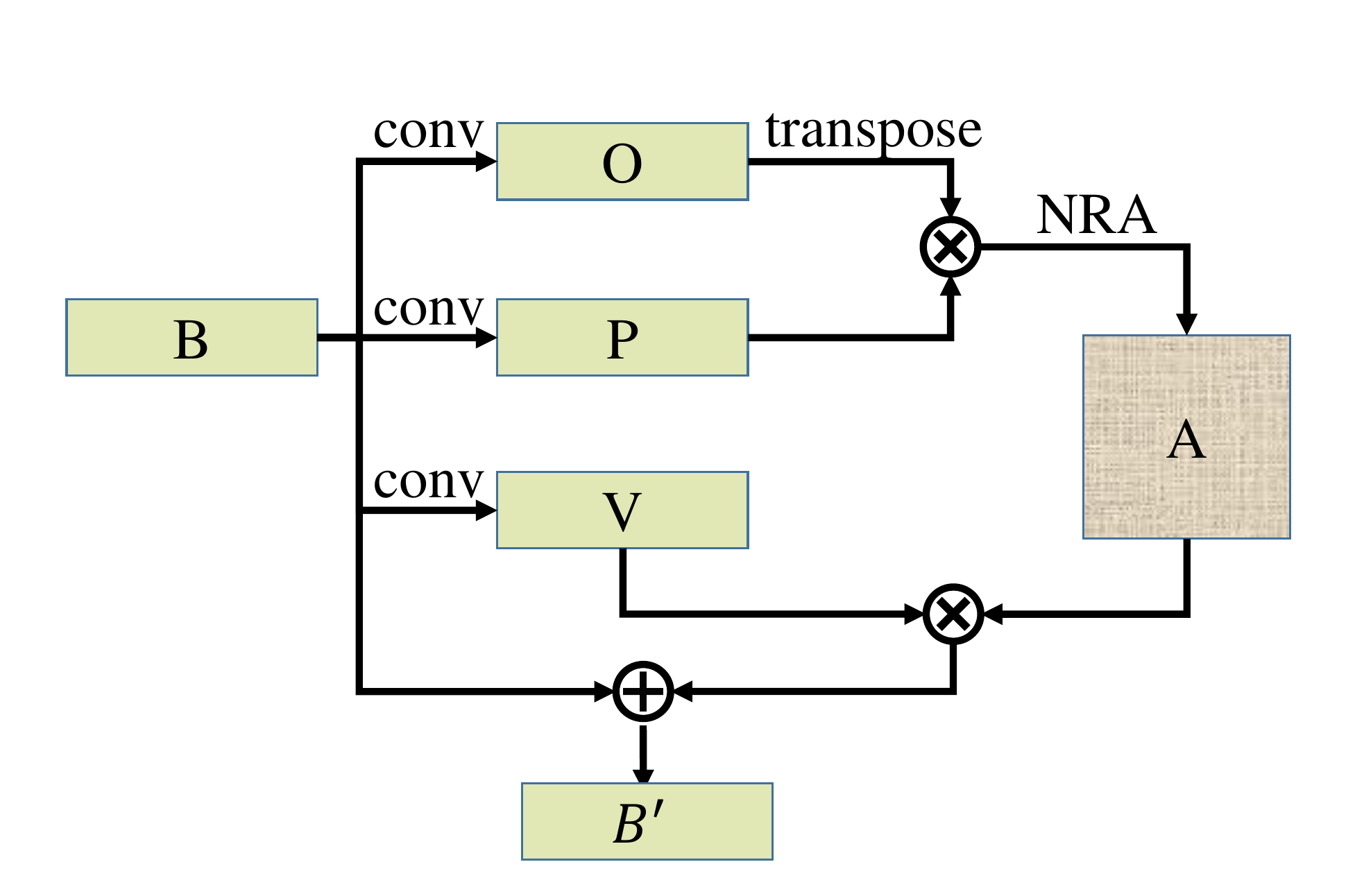}
\end{center}
   \caption{llustration of the proposed Position-independent attention module. $\oplus$ represents the element-wise sum operation. And $\otimes$ means matrix multiplication operation.}
\label{piam}
\end{figure}

The self-attention mechanism has a talent for capturing the internal correlation of features, which is then used to update the original features. However, the previous self-attention module is usually designed for regular feature maps, and the proposed PCM and RCM put forward requirements for modeling the correlation between irregular feature set. Here we proposes a PiAM for feature correlation modeling and enhancement of irregular feature set.

As the structure shown in Figure \ref{piam}, given an input feature set $B \in \mathbb{R}^{C \times K}$, where $C$ and $K$ represent the number of feature channels and set length respectively, we first apply two different convolution layers to generate two feature maps $O$ and $P$, in which $\{O,P\} \in \mathbb{R}^{\frac{C}{4} \times K}$. Different from the squared difference of Euclidean distance, we calculate the correlation coefficient between any elements of the feature set through matrix multiplication operation:
\begin{equation}
A_{i,j} = O_{:, i} \times P_{:, j},
\end{equation}
\noindent where $A \in \mathbb{R}^{K \times K}$ represents the correlation coefficient matrix, and $A_{i, j}$ represents the correlation coefficient between elements $i$ and $j$. Subsequently, we perform normalizing rank aggregation (NRA) on matrix $A$:
\begin{equation}
A_{i,j} = \frac{A_{i,j}}{\sum{A_{i,:}}},
\end{equation}

\noindent It ensures that the enhanced feature statistics will not change significantly, because the sum of the weight of different feature elements is 1. The feature set will then be updated through matrix multiplication between $B$ and $A$ to enhance feature representation. At last, the updated  feature set is multiplied by a scale parameter $\alpha$ and performs a element-wise sum operation with the original feature set $B$ to acquire the final output $B' \in \mathbb{R}^{C \times K}$, where $\alpha$ is initialized to 0 and can be used to confirm the stability at the beginning of training.

In general, our proposed PiAM can perform feature similarity measurement and enhancement on irregular sets. Compared with the previous attention module for regular feature maps, it can effectively deal with irregular feature sets. Moreover, by ignoring the two square terms in the Euclidean distance expansion term and directly calculating the correlation coefficient matrix, the calculation cost can be significantly reduced.

\subsection{Loss function}
Considering the large variation in the number of pixels in each class in the training set, we adopt weighted cross-entropy loss function $L_{context}$ to train proposed model:

\begin{equation}
L_{context}(y_i, p_i)=\sum_{i=1}^{N}{-w_iy_ilog(p_i)},
\end{equation}

\noindent where $y_i$ represent the ground truth of current pixel and $p_i$ is the prediction result by softmax. $w_i$ represents the weight of the $i_{th}$ class, which is calculated through the median frequency balance \cite{median}:

\begin{equation}
w_i = f_{median}/f_i,
\end{equation}

\begin{equation}
f_i = n_i / \sum_{i=1}^{N}{n_i},
\end{equation}

\noindent where $f_{median}$ is the median of all these frequencies.
In addition, we introduce an auxiliary loss $L_{prior}$ to the output of multi-scale guided pre-segmentation module in Figure \ref{overall_structure} based on above weighted cross-entropy loss. And the total loss is denoted as:
\begin{equation}
L = L_{context} + \lambda L_{prior}
\end{equation}
\noindent in which $\lambda$ is a hyperparameter used to control the weight between $L_{context}$ and $L_{prior}$.

\section{Experiments}
\begin{figure}[t]
\setlength{\abovecaptionskip}{0.cm}
\setlength{\belowcaptionskip}{-0.cm}
\begin{center}
   \includegraphics[width=8cm]{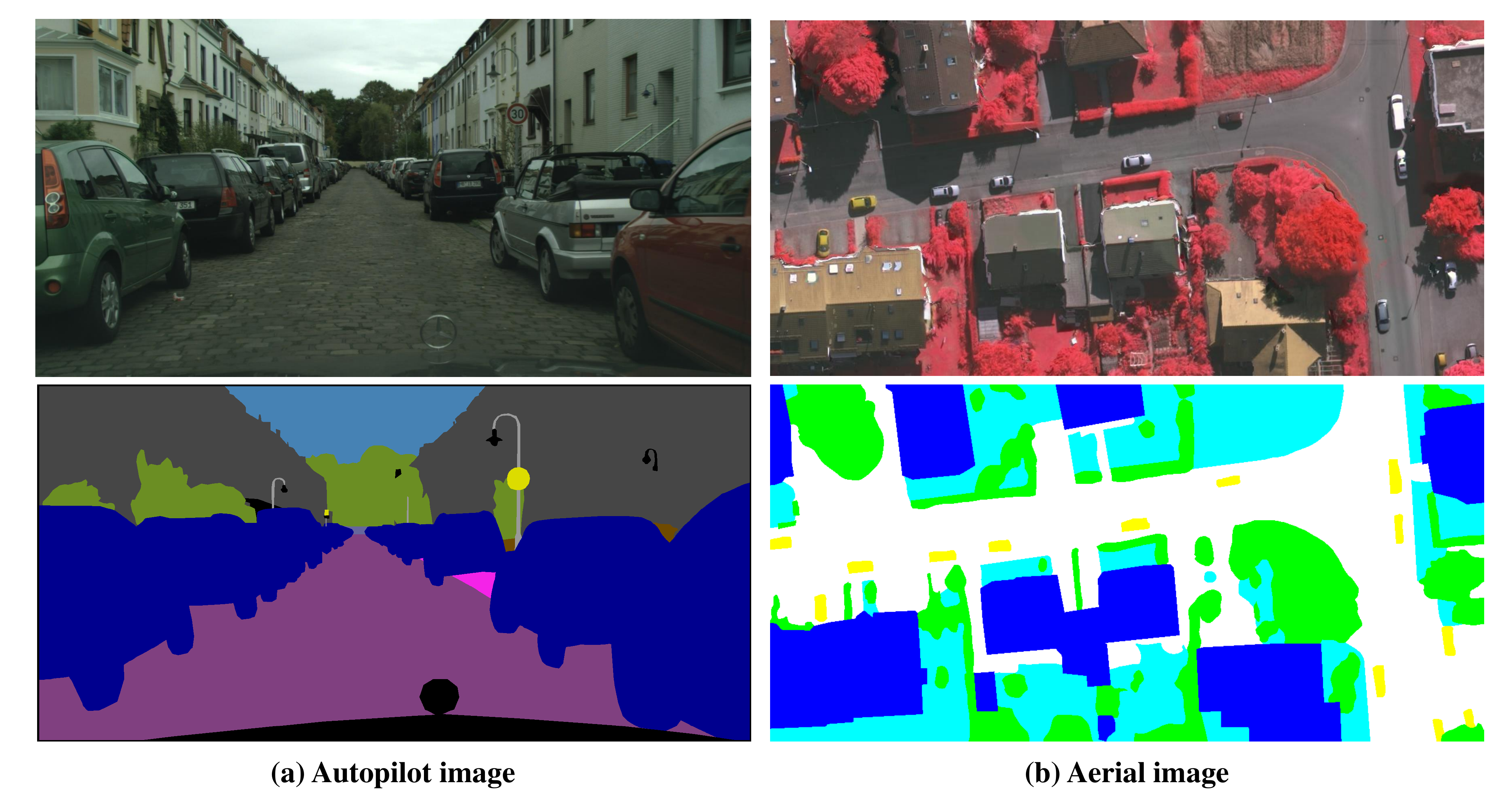}
\end{center}
   \caption{Illustration of the difference between autopilot images and aerial images. Due to the different imaging views (front-view and top-view), there are great differences in the distribution and scale of objects.}
\label{contrast}
\end{figure}

To evaluate our proposed method, we conduct extensive experiments on Cityscapes dataset \cite{cityscapes} and ISPRS Vaihingen dataset \cite{vaihingen}, which differ greatly in the spatial distribution and the scale of objects. 
The spatial distribution is related to the imaging perspective, the former is the front view and the latter is the top view. The scale variation depends on the distance between the object and the camera and the size of the object itself, where the distance between the objects and the camera in aerial image is approximately equal. For instance, the size of the car varies greatly in autopilot images, but roughly the same in aerial images, as shown in Figure \ref{contrast}. And the most important is that they are the authoritative benchmarks in the field of autonomous driving and remote sensing for semantic segmentation, respectively.

\subsection{Experimental Setup}
\subsubsection{Datasets}
\noindent
\textbf{Cityscapes.} 
This is a large-scale dataset used for semantic urban scene understanding, which contains 5,000 images with fine annotations and 20,000 images with coarse annotations. This dataset is collected from 50 different cities and includes a total of 30 classes, 19 of which are used for actual training and validation. It is noted that in our experiments, we only use 5,000 images with fine annotations as our dataset, which is divided into 2,975, 500 and 1,525 images for training, validation and online testing.

\noindent
\textbf{ISPRS Vaihingen.} 
This dataset consists of 33 airborne tiles of Vaihingen, whose size is about 2500 $\times$ 2000 pixels. Each of them contains a high-resolution TOP (True Ortho Photo) tile and corresponding DSM (Digital Surface Model) and nDSM (normalized Digital Surface Model) data with a GSD (Ground Sampling Distance) of 9 $cm$. The TOP file contains three bands corresponding to IR (near-infrared), R (red) and G (green) bands respectively. Among these images, 16 tiles are used for training, in which all pixels are classified as impervious surface, building, low vegetation, tree, car and background. The remaining 17 tiles are withheld for testing. Note that we only adopt IRRG images without DSM and nDSM data during the process of training.

\subsubsection{Implementation Details}
We implement our method in Pytorch. Following \cite{PSPNet, DeepLabV3}, we initialize the learning rate to 0.01 and adopt the poly learning rate policy whose learning rate is updated by $\left ( 1-\frac{iter}{total\_iter} \right )^{0.9}$ after each iteration. For the optimizer, we use stochastic gradient descent (SGD) with weight decay 0.0005 and momentum 0.9. To ensure the stability of parameters in normalization layers, Synchronized BN \cite{syncbn} is adopted to collect the statistics of batch normalization on the whole mini-batch. Specifically, all experiments are trained for 200 epochs with batch size 8 on 4 Tesla V100 GPUs with 16G memory per GPU. To avoid overfitting, we employ common data augmentation strategies, including random scaling in the range of [0.5, 2], random horizontal flipping, and random cropping. In particular, rotating at 90$^\circ$ interval is employed for ISPRS Vaihingen dataset to simulate the changes in flight direction. We set the crop size to 512 $\times$ 1024 for Cityscapes dataset and 768 $\times$ 768 for ISPRS Vaihingen dataset. As for the loss function, the weight $\lambda$ of the prior loss for multi-scale guided pre-segmentation is set to 0.8.

\subsubsection{Standard Pixel-wise Evaluation Metrics}
\noindent
\textbf{Cityscapes.} 
To assess performance, Cityscapes benchmark relies on the standard Jaccard Index, commonly known as intersection-over-union (IoU):
\begin{equation}
IoU = TP ⁄ (TP+FP+FN),
\end{equation}
\noindent where TP, FP, and FN are the numbers of true positive, false positive, and false negative pixels, respectively, determined over the whole test set.

~\\
\noindent
\textbf{ISPRS Vaihingen.} 
Following the evaluation metrics of ISPRS 2D Semantic Labeling Challenge \cite{vaihingen}, per-class $F_{1}$ score and overall accuracy (OA) are adopted to evaluate the performances of our proposed model. F1 score and OA are defined as follows:
\begin{equation}
\footnotesize
F_{1}=\frac{2TP}{2TP+FP+FN},~~~~
OA=\frac{TP+TN}{TP+FP+TN+FN},
\end{equation}

\subsubsection{Scale-sensitive IoU (S-IoU)}

\begin{figure}[t]
\setlength{\abovecaptionskip}{0.cm}
\setlength{\belowcaptionskip}{-0.cm}
\begin{center}
   \includegraphics[width=8cm]{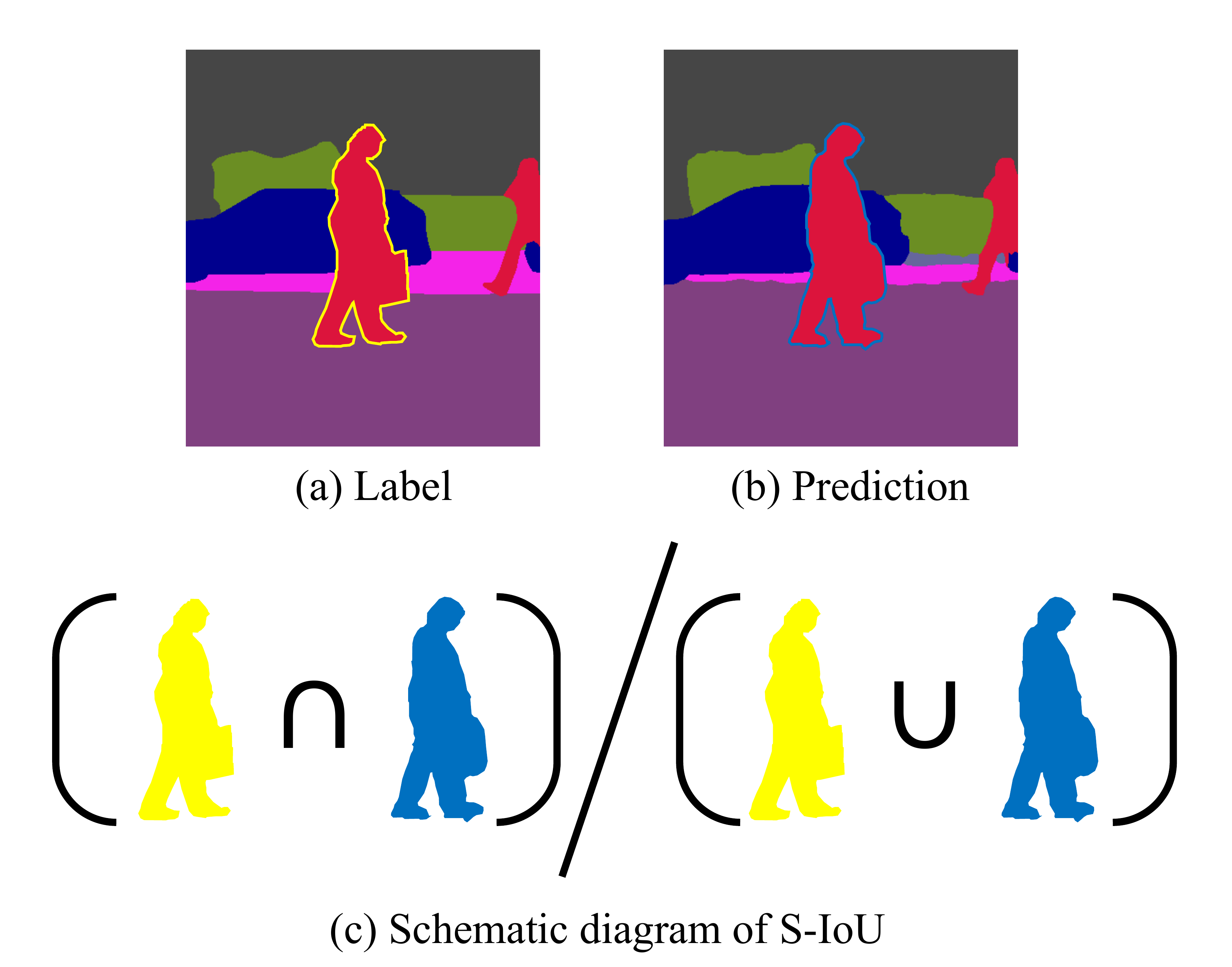}
\end{center}
   \caption{Illustration of S-IoU used for object evaluation at different scales.}
\label{siouyl}
\end{figure}

However, the above standard evaluation metrics regard pixel as the basic evaluation elements, and then count the pixel-wise performance of each class separately. This method can only reflect the performance of the model in each class, and cannot evaluate the performance of the model against multi-scale objects. To evaluate the performance of models at various scale objects, previous works \cite{gao2019focusnet:, gscnn} usually qualitatively divide objects of the same class into specific scales, for example, it was intuitively believed that buses belong to objects with large size while bicycles belong to small objects. Since the size of the objects varies greatly with the distance to camera in natural images, these methods cannot quantitatively reflect the performance of the model for multi-scale objects.

Thus, we propose an evaluation metric called scale-sensitive IoU (S-IoU) for quantitative evaluation. It regards each instance object as the basic evaluation unit with the help of the ground truth of the instance segmentation task. 

Specifically, for each instance object, we first calculate its scale (area) and then match its mask in the prediction result. Subsequently, the intersection-over-union between the predicted mask and the label mask is calculated:
\begin{equation}
S-IoU=\frac{area(label) \cap area(prediction)}{area(label) \cup area(prediction)},
\end{equation}

\subsection{ABLATION study}

\begin{figure*}[htbp]
\centering
\includegraphics[width=17cm]{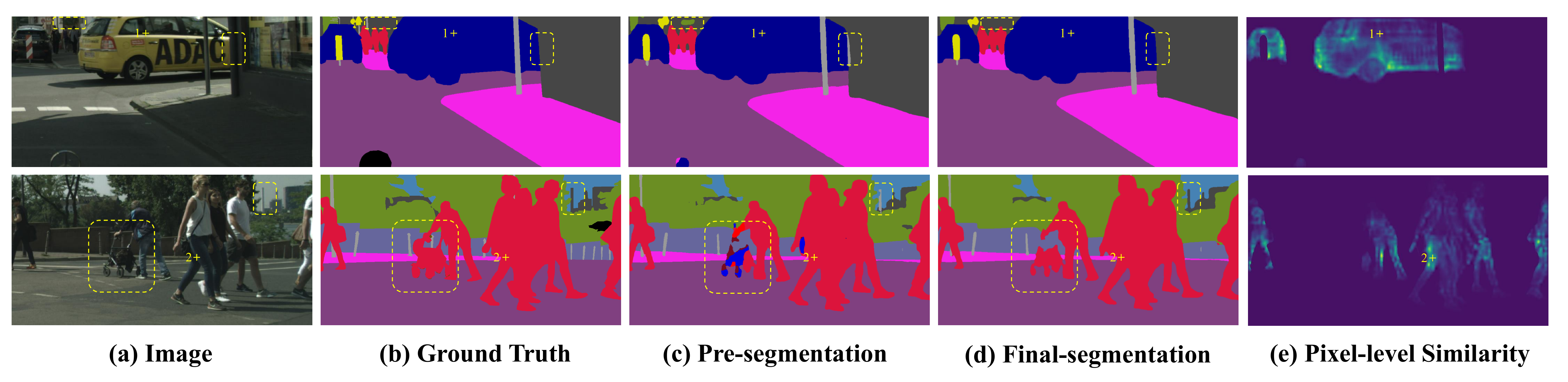}
\caption{Visualization results of context modules on Cityscapes val set. For each example, we show the input image, ground truth, pre-segmentation result, final segmentation result and intra-class similarity map correspondence to the markers in the input image.}
\label{context_visual}
\end{figure*}

\begin{figure}
\centering
\includegraphics[width=8cm]{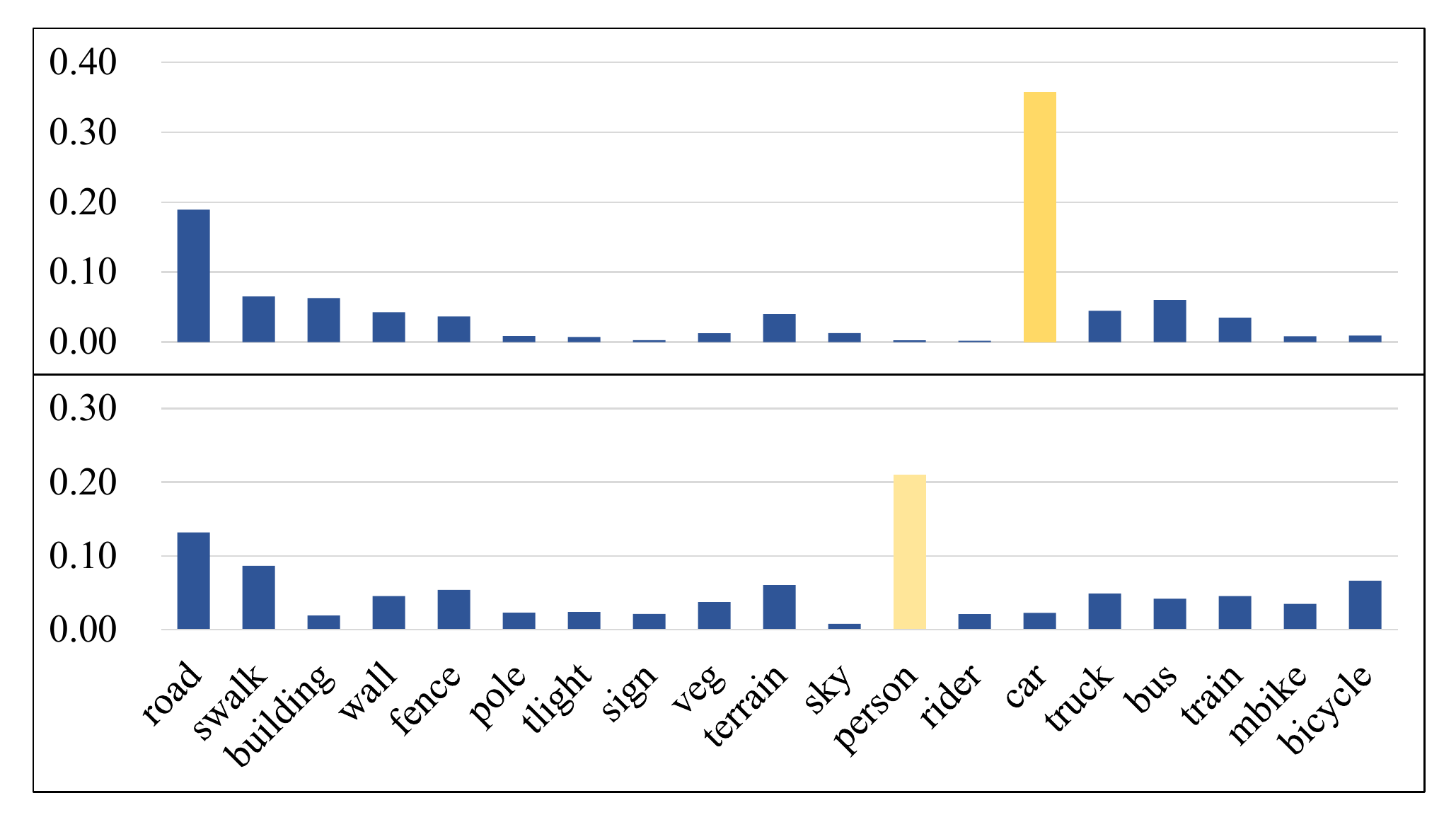}
\caption{Visualization results of class-based region dependencies.}
\label{inter_class}
\end{figure}

In our network, PCM and RCM are used to hierarchically capture global context information from pixel-level and region-level. In this section, we show the results comparing to our baseline dilated ResNet-101 in terms of standard mIoU.

\begin{table}
\linespread{0.8} 
\begin{center}
\setlength{\tabcolsep}{5pt}
\begin{tabular}{lcccc}
\toprule[0.5pt]
Method  & PCM & RCM& mIoU (\%) \\

\hline \hline

Baseline  &         &      &   75.37 \\
HCNet  &        \checkmark &         &      78.59 \\
HCNet  &         &       \checkmark &      77.26 \\
HCNet  &        \checkmark &        \checkmark &      79.86 \\
\bottomrule[0.5pt]
\end{tabular}
\caption{Ablation study for our proposed hierarchical context modules on Cityscapes val set. For fair comparison, both baseline and HCNet use ResNet-101 as the backbone.}
\label{Ablation}
\end{center}
\end{table}

In Table \ref{Ablation}, we evaluate the effectiveness of our proposed context modules on the Cityscapes validation set. It can be seen that employing the PCM individually yields a result of 78.59\% in mIoU, which outperforms baseline by 3.22\%. Meanwhile, HCNet with RCM individually realizes 77.26\% in mIoU, which results in a 1.89\% improvement compared to baseline. 
It can also be seen that the RCM has achieved fewer improvements than the PCM. This is because only capturing region-level context information makes the model lack pixel-level detailed information. 
After integrating pixel-level and region-level context, the performance of our hierarchical context network improves to 79.86\%  as expected.

To better understand the context modules visually, we provide visualization results for our hierarchical context modules. As shown in Figure \ref{context_visual}, we also visualize the pre-segmentation and final segmentation results in columns 3 and 4, with the yellow dashed boxes marking the improved areas. For PCM, we select one marker on each image and visualize the similarities with all the other positions belonging to the same homogeneous region as a similarity map of size $H \times W$ in column 5. Specially, the similarities between positions belonging to different regions are filled with zero according to the pre-segmentation results. For example, the first marker in the image (1) is pre-segmented as a car and the corresponding highlighted areas on the similarity map also belong to the car. Similarly, the second marker is pre-segmented into person, and its corresponding similarity map is only responded in areas of the corresponding region. For RCM, we can obtain the similarities of a certain class-based region with all class-based regions with shape of $1 \times N$. Corresponding to the two markers in Figure \ref{context_visual}, we visualize the correlations of car and person as a histogram in Figure \ref{inter_class}. It can be seen that these are specific correlations between different class-based regions, which can further enhance the discriminability of features.

We further exhibit the comparison of the increased FLOPs and GPU memory of HCNet and self-attention compared to baseline (dilated ResNet-101). As shown in Table \ref{flops}, HCNet achieves a 1.03\% improvement in mIoU compared to self-attention, but significantly reduces FLOPs and GPU memory by about 40\% and 85\%, demonstrating that our approach can capture global context information in a more sparse and effective way.

\begin{table}
\linespread{1.25} 
\begin{center}
\setlength{\tabcolsep}{5pt}
\begin{tabular}{lrrc}
\toprule[0.5pt]
Method &  GFLOPs ($\blacktriangle$) & Memory(M $\blacktriangle$) & mIoU \% \\
\hline\hline
Baseline & 0.00  & 0.00   & 75.37\\
self-attention & 121.62   &  471.04 & 78.83\\
HCNet &71.63 &  70.08 & 79.86 \\
\bottomrule[0.5pt]
\end{tabular}
\caption{Comparison of self-attention and HCNet. FLOPs and Memory usage are estimated for an input of $1 \times 3 \times 512\times 1024$.}
\label{flops}
\end{center}
\end{table}

\subsection{Scale-sensitive Evaluation}

\begin{table*}[t]
\linespread{0.8} 
\begin{center}
\setlength{\tabcolsep}{5pt}
\begin{tabular}{lcccc}
\toprule[0.5pt]
Method   & \begin{tabular}[c]{@{}c@{}}mS-IoU\\ area $\in$ {[}0, 2500) \end{tabular} & \begin{tabular}[c]{@{}c@{}}mS-IoU\\ area $\in$ {[}2500, 62500) \end{tabular} & \begin{tabular}[c]{@{}c@{}}mS-IoU\\ area $\in$ {[}62500, $\infty$)\end{tabular} & mS-IoU \\
\hline\hline
Baseline & 49.32  & 74.51   & 87.43 & 62.41\\
PSPNet & 51.01 & 75.26 & 89.98 & 65.07 \\
DANet & 53.47&	75.29&	89.06&	66.43\\
HCNet &54.17&	75.80&	90.82&	67.78 \\
\bottomrule[0.5pt]
\end{tabular}
\caption{Comparison in terms of S-IoU vs state-of-the-art models on the Cityscapes val set. It can be seen that HCNet achieves the best interval and overall performance compared to the other three methods.}
\label{siou}
\end{center}
\end{table*}

In Table \ref{siou}, we compare the performance of our HCNet with Baseline, DANet and PSPNet using our proposed S-IoU. Specifically, we first calculate the scale (measured by area) and S-IoU of all instance objects. Whereafter, all the objects are divided into three scale intervals by area (number of pixels) as shown in Table \ref{siou}. For the sake of comparison, we calculate the interval evaluation metric mS-IoU by averaging the S-IoU of all objects in each interval. Similarly, the overall evaluation metric can also be obtained by averaging the S-IoU of all objects. It can be seen that HCNet achieves the best interval and overall performance compared to the other three methods. This confirms that our proposed hierarchical context modules are extremely effective  for identifying multi-scale objects.

\begin{table*}[t]
\setlength{\abovecaptionskip}{3pt}
\setlength{\belowcaptionskip}{-3pt}
\linespread{0.8} 
\begin{center}
\setlength{\tabcolsep}{3pt}

\small
\begin{tabular}{l|c|ccccccccccccccccccc}
%\hline
Methods	& \rotatebox{90}{mIoU}	& \rotatebox{90}{road}	& \rotatebox{90}{sidewalk}	& \rotatebox{90}{building}	&\rotatebox{90}{wall}	&\rotatebox{90}{fence}	&\rotatebox{90}{pole}	&\rotatebox{90}{traffic light}	&\rotatebox{90}{traffic sign}	&\rotatebox{90}{vegetation}	&\rotatebox{90}{terrain}&	\rotatebox{90}{sky}	&\rotatebox{90}{person}	&\rotatebox{90}{rider}	&\rotatebox{90}{car}	& \rotatebox{90}{truck}&	\rotatebox{90}{bus}	&\rotatebox{90}{train}&	\rotatebox{90}{motorcycle}	&\rotatebox{90}{bicycle} \\

\hline\hline

DeepLab-v2 \cite{deeplabv2:}	& 70.4 &	97.9 &	81.3 &	90.3 &	48.8 &	47.4 &	49.6 &	57.9 	&67.3 &	91.9 &	69.4& 	94.2 &	79.8 &	59.8 	&93.7 &	56.5 	&67.5 &	57.5 &	57.7 &	68.8\\

RefineNet \cite{RefineNet}	&73.6 	&98.2 	&83.3 	&91.3 	&47.8 	&50.4 	&56.1 	&66.9 	&71.3 	&92.3 	&70.3& 	94.8 	&80.9 	&63.3 	&94.5 	&64.6 	&76.1 	&64.3 	&62.2 	&70.0 \\

%DUC \cite{DUC}	&77.6 	&98.5 	&85.5 	&92.8 	&58.6 	&55.5 	&65.0 	&73.5 	&77.9 	&93.3 	&72.0 	&95.2 	&84.8 	&68.5 	&95.4 	&70.9 	&78.8 	&68.7 	&65.9 	&73.8 \\

%ResNet-38 \cite{Resnet38}	&78.4 	&98.5 	&85.7 	&93.1 	&55.5 	&59.1 	&67.1 	&74.8 	&78.7 	&93.7 	&72.6 	&95.5 	&86.6 	&69.2 	&95.7 	&64.5 	&78.8 	&74.1 	&69.0 	&76.7 \\

%BiSeNet \cite{BiSeNet}	&78.9 	&-	&-	&-	&-	&-	&-	&-	&-	&-	&-	&-	&-	&-	&-	&-	&-	&-	&-	&- \\

%DFN \cite{DFN}	&79.3 	&-	&-	&-	&-	&-	&-	&-	&-	&-	&-	&-	&-	&-	&-	&-	&-	&-	&-	&- \\

%PSANet \cite{PSANet}	&80.1 	&-	&-	&-	&-	&-	&-	&-	&-	&-	&-	&-	&-	&-	&-	&-	&-	&-	&-	&- \\

DenseASPP \cite{DenseASPP}	&80.6 	&98.7 	&87.1 	&93.4 	&60.7 	&62.7 	&65.6 	&74.6 	&78.5 	&93.6 	&72.5& 	95.4 	&86.2 	&71.9 	&96.0 	&78.0 	&90.3 	&80.7 	&69.7 	&76.8 \\

%SVCNet \cite{SVCNet}	&81.0 	&-	&-	&-	&-	&-	&-	&-	&-	&-	&-	&-	&-	&-	&-	&-	&-	&-	&-	&-\\

BFP \cite{bfp} &81.4 &98.7 &87.0 &93.5 &59.8 &63.4 &68.9 &76.8 &80.9 &93.7 &72.8 &95.5 &87.0 &72.1 &96.0 &77.6 &89.0 &86.9 &69.2 &77.6 \\

DANet \cite{DANet}	&81.5 	&98.6 	&86.1 	&93.5 	&56.1 	&63.3 	&69.7 	&77.3 	&81.3 	&93.9 	&72.9 	&95.7 	&87.3 	&72.9 	&96.2 	&76.8 	&89.4 	&86.5 	&72.2 	&78.2 \\

%HRNet \cite{hrnet}	&81.6 	&-	&-	&-	&-	&-	&-	&-	&-	&-	&-	&-	&-	&-	&-	&-	&-	&-	&-	&- \\

ACFNet \cite{acfnet}  &81.8 &98.7 &87.1 &93.9 &60.2 &63.9 &71.1 &78.6 &81.5 &94.0 &72.9 &95.9 &88.1 &74.1 &96.5 &76.6 &89.3 &81.5 &72.1 &79.2 \\
CCNet \cite{PSANet}	&81.9 	&-	&-	&-	&-	&-	&-	&-	&-	&-	&-	&-	&-	&-	&-	&-	&-	&-	&-	&- \\
%SPNet \cite{strip}	&82.0 	&-	&-	&-	&-	&-	&-	&-	&-	&-	&-	&-	&-	&-	&-	&-	&-	&-	&-	&-\\

HCNet	&82.8 	&98.6 	&87.5 	&94.1 	&61.4 	&64.9 	&73.0 	&78.8 	&82.1 	&93.9 	&74.0 	&95.7 	&88.3 	&73.6 	&96.2 	&80.0 	&91.8 	&87.6 	&71.6 	&79.7 \\ 

\bottomrule[0.5pt]
\end{tabular}
\end{center}
\caption{Comparison with state-of-the-art on Cityscapes test set. For fair comparison, all methods only adopt data with fine annotations for training. HCNet outperforms existing approaches and achieves 82.8\% in Mean IoU.}
\label{city_table}
\end{table*}

\subsection{Benchmark Evaluation}
To get optimal performance on the benchmark test set, we use our best model (i.e., HCNet with multi-scale guided pre-segmentation and hierarchical context modules). Additionally, we use multi-scale inference schemes with scales 0.5, 1.0, and 2.0.

\subsubsection{Cityscapes Benchmark}
In Table \ref{city_table}, we compare against published methods on the Cityscapes test set without using the coarse data. Among these methods, DeepLab-v2, RefineNet and DenseASPP, which utilize multi-scale context aggregation, have not achieved leading performance due to the lack of adaptive feature aggregation for each position. However, DANet, ACFNet, and CCNet using self-attention mechanisms generally outperform multi-scale methods, which reflects the necessity of adaptive modeling global context information. By combining the PCM and the RCM, HCNet can model global context information in a sparse manner and effectively capture multi-granularity features. Experimental results show that Our model achieves the best performance 82.8\%, which is extremely competitive with recent state-of-the-art models. It is also important to stress that our model outperforms very strong to other state-of-the-art models in classes with large object size (such as bus, train, and wall) and small object size (such as pole and traffic light).

\subsubsection{ISPRS Vaihingen Benchmark}
We carry out experiments on ISPRS Vaihingen Benchmark to further evaluate the effectiveness of our method. Table \ref{vaihingen_table} shows the quantitative results of our HCNet on ISPRS Vaihingen test set. ADL\_3 \cite{ADL_3}, which adopts CNN and uses some post-processing schemes such as conditional random field for classification, obtains an overall accuracy of 88.0\%. DST\_2  \cite{DST_2} using the FCN variant and ONE\_7 \cite{ONE_7} using the SegNet variant are limited by the local receptive field, achieving overall accuracy of 89.1\% and 89.8\% respectively. The most powerful competitors are GSN \cite{GSN} using gated convolution and DLR\_10 \cite{DLR_10} combined with boundary detection. Although we did not use auxiliary techniques such as conditional random fields and boundary detection, HCNet ranks $1^{st}$ both in per-class $F_{1}$ score and overall accuracy, compared with other state-of-the-art methods. In particular, our approach achieves 88.6\% in car class (average size $38\times 18$ pixels), outperforming the previous best model by a large margin. This is because our proposed HCNet can simultaneously model global context and capture multi-granularity features.

\begin{table}
\setlength{\abovecaptionskip}{3pt}
\setlength{\belowcaptionskip}{15pt}
\linespread{0.8} 
\begin{center}
\setlength{\tabcolsep}{3pt}
\small
\begin{tabular}{lcccccc}
\toprule[0.5pt]

Method & Imp surf  & Building & Low veg	&Tree& 	Car & OA  \\
\hline\hline
 ADL\_3   &  89.5 &93.2 &82.3   &88.2       &63.3 & 88.0\\
DST\_2  & 90.5    &93.7    &83.4   & 89.2 &  72.6    & 89.1\\
ONE\_7  &91.0    & 94.5  & 84.4  & 89.9 & 77.8     & 89.8\\
DLR\_10  & 92.3     & 95.2  & 84.1 & 90.0 & 79.3     & 90.3\\
GSN  & 92.2     & 95.1  & 83.7 & 89.9 &82.4      & 90.3\\
HCNet  & 93.1     & 96.1  & 84.9 & 90.1 & 88.6     & 91.4\\
\bottomrule[0.5pt]
\end{tabular}
\end{center}
\caption{Quantitative comparisons between our method and other related methods on ISPRS Vaihingen test set.}
\label{vaihingen_table}
\end{table}

\begin{figure*}[htbp]
\centering
\includegraphics[width=17cm]{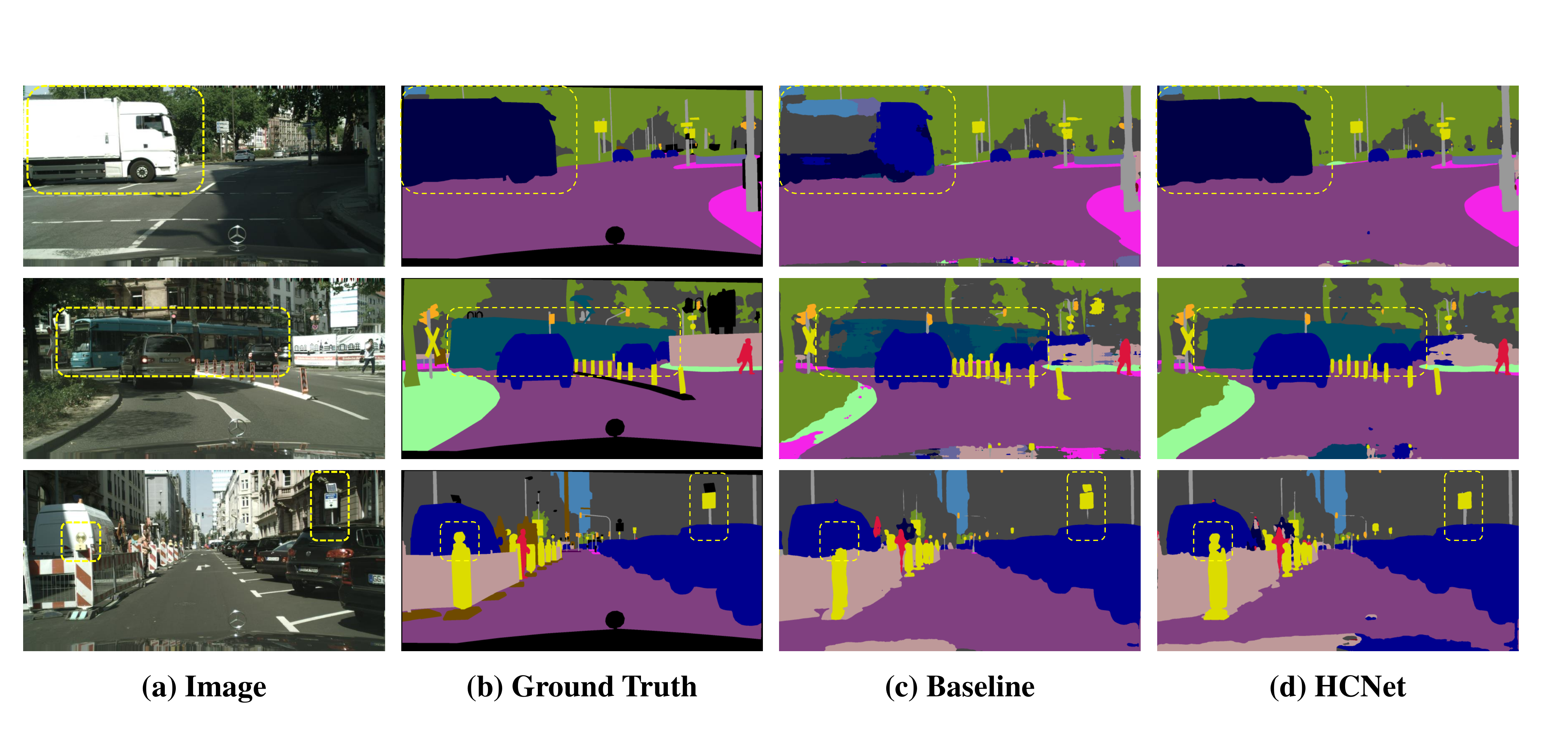}
\caption{Visualization results of HCNet on Cityscapes val set. The yellow dashed box marks the challenging areas. In addition, the black area in the ground truth is the unlabeled class, which is not included in the final evaluation.}
\label{city_visual}
\end{figure*}

\begin{figure}
\setlength{\abovecaptionskip}{3pt}
\setlength{\belowcaptionskip}{-3pt}
\centering
\includegraphics[width=8cm]{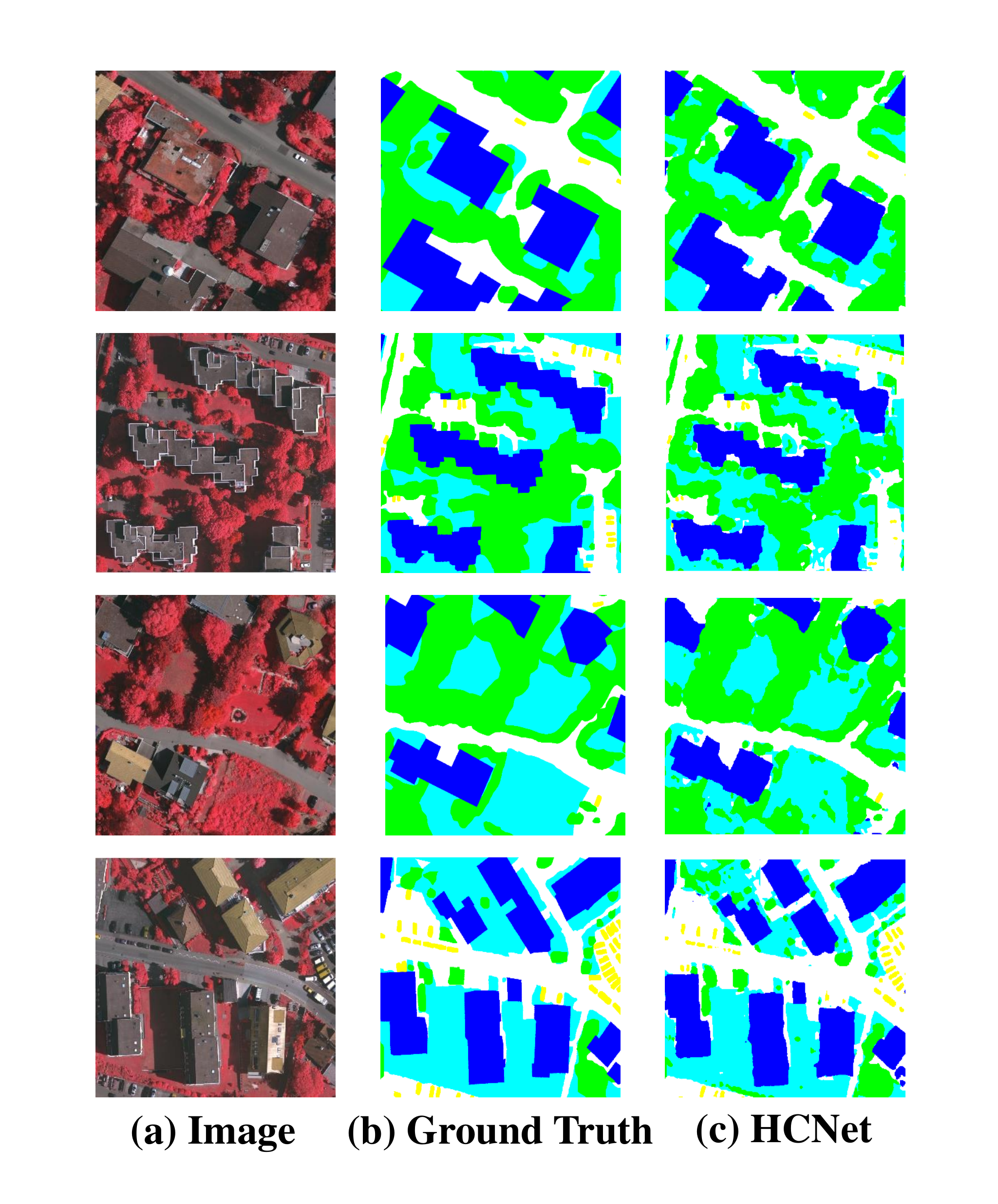}
 \caption{Visual results on ISPRS Vaihingen val set, where the impervious surface, building, tree, low vegetation and car are marked as white, blue, green, cyan, and yellow respectively.}
\label{vaihingen}
\end{figure}

\subsection{qualitative results}
In Figure \ref{city_visual}, we provide the qualitative comparison between the results of baseline and our HCNet on Cityscapes validation set. As shown in the yellow dashed boxes of the first example, baseline cannot identify truck considering of local receptive field, while our HCNet makes relatively accurate predictions due to the rich contextual information capturing through the hierarchical context modules. Furthermore, the scene in the second line has a highly heterogeneous appearance, which leads to a misjudgment of baseline. Thanks to the global context information provided by PCM and RCM, HCNet is able to effectively enhance the discriminability of features for accurate reasoning and prediction. In particular, in the third line, we can see that our model also has good performance for small objects.

Figure \ref{vaihingen} illustrates a few examples of segmentation results on ISPRS Vaihingen test set. It can be seen that HCNet can accurately segment objects with great differences in appearance (such as buildings) and small-scale objects (such as cars), which is benefited from our proposed context modules that can effectively capture contextual dependencies and multi-scale features.

\section{Conclusion}
In this paper, we propose a hierarchical context network (HCNet) for semantic segmentation, which can capture global context information more sparsely than the self-attention mechanism. Specifically, based on the region partition prior generated by proposed multi-scale guided pre-segmentation, PCM and RCM are designed to hierarchically model global context dependencies from pixels-level and region-level. Meanwhile, through aggregating fine-grained pixel context features and coarse-grained region context features, HCNet can harvest multi-granularity representations to more robustly identify multi-scale objects. Extensive experiments conducted on Cityscapes and ISPRS Vaihingen dataset prove the effectiveness of our proposed modules. Furthermore, our method also achieves the optimal multi-scale performance, which is evaluated through our proposed scale-sensitive IoU.

{\small
\bibliographystyle{ieee}
\bibliography{egbib}
}

\end{document}